\documentclass[letterpaper]{article}
\usepackage{aaai17}
\usepackage{color}
\usepackage{times}
\usepackage{helvet}
\usepackage{courier}
\usepackage{graphicx} % more modern
\usepackage{subfigure}
\usepackage{epstopdf}
\usepackage{multirow}
\usepackage{algorithm,algorithmic,amsmath,amsthm,amsfonts,bbm, dsfont}
\usepackage{verbatim,graphicx,subfigure}
\usepackage{mathtools}
\usepackage{algorithm}
\usepackage{algorithmic}

\usepackage{bm,amssymb}

\newcommand{\G}{\mathcal{G}}
\newcommand{\A}{\mathcal{A}}

\newcommand{\R}{\mathbb{R}}

\newcommand{\vct}[1]{\bm{#1}}
\newcommand{\mtx}[1]{\bm{#1}}
\newcommand{\vx}{\vct{x}}

\newcommand{\vy}{\vct{y}}
\newcommand{\vz}{\vct{z}}

\newcommand{\va}{\vct{a}}

\newcommand{\vw}{\vct{w}}
\newcommand{\vwstar}{\vct{w}_{\star}}

\newcommand{\vc}{\vct{c}}

\newcommand{\mX}{\mtx{X}}

\newcommand{\T}{\mathcal{T}}

\newcommand{\gstar}{g_{\star}}
\newcommand{\Phistar}{\Phi_{\star}}

\newcommand{\bi}{\{ i \}}
\def\T0{T_0}

\def\bbE{\mathbb{E}}
\def\cG{\mathcal{G}}
\def\bbR{\mathbb{R}}
\newcommand{\defeq}{\mbox{$\;\stackrel{\mbox{\tiny\rm def}}{=}\;$}}
\newcommand{\gensim}{\text{CSI }}
\newcommand{\stocgensim}{\text{Stochastic \gensim}}
\newtheorem{conjecture}{Conjecture}
\def\reals{{\mathbb R}}

\def\expect{{\mathbb E}}

\newcommand{\beq}{\begin{equation}}
\newcommand{\eeq}{\end{equation}}

\begin{document}
\title{On Learning High Dimensional Structured Single Index Models}
\author{Ravi Ganti$^1$, Nikhil Rao$^2$\thanks{The first two authors contributed equally to this paper.}, Laura Balzano$^3$, Rebecca Willett$^4$, Robert Nowak$^4$\\
$^1$\ Walmart Labs, San Bruno, CA\\
gmravi2003@gmail.com\\
$^2$\ Technicolor Research and Innovation, Los Altos, CA\\
$^3$\ EECS, University of Michigan, Ann Arbor, MI\\
$^4$\ ECE, University of Wisconsin, Madison, WI\\}
\maketitle

\begin{abstract}
Single Index Models (SIMs) are simple yet flexible semi-parametric models for machine learning, where the response variable is modeled as a monotonic function of a linear combination of features. Estimation in this context requires learning both the feature weights and the nonlinear function that relates features to observations. While methods have been described to learn SIMs in the low dimensional regime, a method that can efficiently learn SIMs in high dimensions, and under general structural assumptions, has not been forthcoming. In this paper, we propose computationally efficient algorithms for SIM inference in high dimensions with structural constraints. Our general approach specializes to sparsity, group sparsity, and low-rank assumptions among others. Experiments show that the proposed method enjoys superior predictive performance when compared to generalized linear models, and achieves results comparable to or better than single layer feedforward neural networks with significantly less computational cost.
\end{abstract}

\section{Introduction}
\label{sec:intro}

High-dimensional machine learning is often tackled using generalized linear models,
where  a response variable $Y \in \reals$ is related to a feature
vector $X \in \reals^d$ via
\beq
\label{eq:sim}
\expect[Y|X=\vx] = \gstar(\vwstar^\top\vx)
\eeq
for some unknown weight vector $\vwstar \in \reals^d$ and some smooth transfer function $\gstar$. Typical examples of $\gstar$ are the logit and probit functions for classification, and the linear function for regression. High dimensional parameter estimation for GLMs has been
widely studied, both from a theoretical and algorithmic point of view~ \cite{van2008high,Mest,park2007l1}. While classical
work on generalized linear models (GLMs) assumes $\gstar$ is known, this function is often unknown in real-world datasets, and hence we need methods that can simultaneously learn both $\gstar$ and  $\vwstar$.

The model in \eqref{eq:sim} with $\gstar$ unknown is called a {\em
  Single Index Model (SIM)} and is a powerful
semi-parametric generalization of a GLM. SIMs were first introduced in the econometrics and statistics literature~\cite{horowitz1996direct,ichimura1993semiparametric,horowitz2009semiparametric}, and have since become popular in  statistical machine learning applications as well.
%Such models are
%particularly popular in
%econometrics~\cite{horowitz2009semiparametric}.
Recently,
computationally and statistically efficient algorithms have been
provided for learning SIMs~\cite{sim_ravi,sim_sham} in low-dimensional
settings where the number of samples/observations $n$ is much larger than the ambient dimension $d$. However, many problems in modern machine learning, signal processing and computational biology are high dimensional, i.e. the number of parameters to learn, $d$ far exceeds the number of data points $n$. For example, in genetics, one has to infer activation weights for thousands of genes with hundreds of measurements.

In this paper, motivated by high-dimensional data analysis problems, we consider learning SIM in high dimensions. This is a hard learning problem because (i) statistical inference is ill-posed, and indeed impossible in the high-dimensional setup without making additional structural assumptions and (ii) unlike GLMs the transfer function itself is unknown and also needs to be learned from the data. To handle these problems we impose additional structure on the unknown weight vector $\vwstar$ which is elegantly captured by the concept of small atomic cardinality~\cite{venkat} and make smoothness assumptions on the transfer function $\gstar$. The concept of small atomic cardinality generalizes commonly imposed structure in high-dimensional statistics such as sparsity, group sparsity, low-rank, %~\cite{hastie2015statistical}
and allows us to design a single algorithm that can learn a SIM with various structural assumptions.

 We provide an efficient algorithm called \gensim (Calibrated Single Index) that can be used to learn SIMs in high dimensions. The algorithm is an optimization procedure that minimizes a loss function that is calibrated to the unknown SIM, for both $\vwstar$ and $\gstar$. CSI alternates between a projected gradient descent step to update its estimate of $\vwstar$ and a function learning procedure called LPAV to learn a monotonic, Lipschitz function.
 We provide extensive experimental evidence that demonstrates the effectiveness of \gensim in a variety of high dimensional machine learning scenarios. Moreover we also show that we are able to obtain competitive, and often better, results when compared to  a single layer neural network, with significantly less computational cost.

%The rest of this paper is organized as follows: after surveying the related literature, we formally set up the problem we are interested in in Section \ref{sec:highd}. In Section \ref{sec:algo}, we introduce our algorithm Calibrated Single Index learning (\gensim) and provide details about implementation and computational considerations. In Section \ref{sec:stoc}, we extend \gensim to a stochastic setting. We compare \gensim to several other methods on a variety of high dimensional structurally constrained problems in Section \ref{sec:expts}, before concluding the paper in Section \ref{sec:conc}.

%In Section~\ref{sec:rw} we survey related work. In Section \ref{sec:highd} we formally set up the problem we are interested in solving and introduce our algorithm in Section \ref{sec:algo}. We compare \gensim to several other methods on a variety of high dimensional structurally constrained problems in Section \ref{sec:expts}, before concluding the paper in Section \ref{sec:conc}
%
%
%
\subsection{Related Work and Our Contributions}
\label{sec:rw}

Alquier and Biau~\cite{alquier2013sparse} consider learning high dimensional single index models. They provide estimators of $\gstar,\vwstar$ using PAC-Bayesian analysis, which relies on reversible jump MCMC, and is slow to converge even for moderately sized problems. \cite{radchenko2015high} learns high dimensional single index models with simple sparsity assumptions on the weight vectors, while \cite{mmc} provide methods to learn SIM's in the matrix factorization setting. While these are first steps towards learning high dimensional SIM's, our method can handle several types of structural assumptions, generalizing these approaches to several other structures in an elegant manner. Restricted versions of the SIM estimation problem with (structured) sparsity constraints have been considered in \cite{plan2014high,rao2014classification}, where  the authors are only interested in accurate parameter estimation and not prediction. Hence, in these works the proposed algorithms do not learn the transfer function. We finally comment that there is also related literature focused on how to query points in order to learn the SIM, such as~\cite{cohen2012capturing}.
%~\cite{radchenko2015high} learns high dimensional single index models with simple sparsity assumptions on the weight vectors; their work and others cited therein use $\ell_1$ or other variable selection techniques.
%{ \bfseries This work requires solving an optimization problem for a grid of regularization parameters, making it intractable for large problems. NR: the sparsity in CSI is also a parameter right?  How about the line below:}

The class of SIM belongs to a larger set of semi-parametric models called multiple index models~\cite{hastie2005elements}, which involves learning a sum of multiple $g_j$ and corresponding $w_j$. Other semi-parametric models ~\cite{friedman1981projection,buja1989linear,ravikumar2009sparse} where the model is a linear combination of functions of the form $g_j(x_j)$ are also popular, but our restrictions on the transfer function allow us to use simple optimization methods to learn $\gstar$.

Finally, neural networks have emerged as a powerful alternative to learn nonlinear transfer functions that can be basically thought of being defined by compositions of nonlinear functions. In the high dimensional setting (data poor regime), it may be hard to estimate all the parameters accurately of a multilayer network, and a thorough comparison is beyond the scope of this paper. Nonetheless, we show that our method enjoys comparable and often superior performance to a single-layer feed forward NN, while being significantly cheaper to train.  These positive results indicate that one could perhaps use our method as a much cheaper alternative to NN in practical data analysis problems, and motivates us to consider ``deep" variants of our method in the future. To the best of our knowledge, simple, practical algorithms with good empirical performance for learning single index models in high dimensions are not available.

%%%%%%%%%%%%%%%%%%%%%%%%%%%%%%%%%%%%%%%%%%%%
%%%%%%%%%%%%%%%%%%%%%%%%%%%%%%%%%%%%%%%%%%%%%%%%%%%%%%%%%%%%%%%%%%%%%%%%%%%%%%%%%%%%%%%%
%%%%%%%%%%%%%%%%%%%%%%%%%%%%%%%%%%%%%%%%%%%%
%%%%%%%%%%%%%%%%%%%%%%%%%%%%%%%%%%%%%%%%%%%%
%%%%%%%%%%%%%%%%%%%%%%%%%%%%%%%%%%%%%%%%%%%%
%%%%%%%%%%%%%%%%%%%%%%%%%%%%%%%%%%%%%%%%%%%%
%%%%%%%%%%%%%%%%%%%%%%%%%%%%%%%%%%%%%%%%%%%%
%%%%%%%%%%%%%%%%%%%%%%%%%%%%%%%%%%%%%%%%%%%%
%%%%%%%%%%%%%%%%%%%%%%%%%%%%%%%%%%%%%%%%%%%%

\section{Structurally Constrained Problems in High Dimensions}
\label{sec:highd}
We now set up notations that we use in the sequel, and set up the problem we are interested to solve.
Assume we are provided i.i.d. data
$\{(\vx_1,y_1),\ldots,(\vx_n,y_n)\}$, where the label $Y$ is generated
according to the model $\bbE[Y|X=x] = \gstar(\vwstar^\top \vx)$ for an unknown
parameter vector $\vwstar \in \reals^d, ~\ n \ll d$ and unknown 1-Lipschitz, monotonic function
$g_\star$.  The monotonicity assumption on $\gstar$ is not unreasonable. In GLMs the transfer function is monotonic. In neural networks the most common activation functions are ReLU, sigmoid, and the hyperbolic tangent functions, all of which are monotonic functions. Moreover, learning monotonic functions is an easier problem than learning general smooth functions, as this learning problem can be cast as a simple quadratic programming problem. This allows us to avoid using costlier non-parametric smoothing techniques such as local polynomial regression~\cite{tsybakov2009introduction}.
We additionally assume that $y\in [-1,1]$ \footnote{We can easily relax this to $y \in [-M,M]$, i.e. bounded is sufficient.}. Let $\mX \in \R^{n \times d}$ be a matrix with each row corresponding to an  $\vx_i$ and let $\vy \in \R^n$ be the corresponding vector of observations. Note that in the case of matrix estimation problems the data $\vx_1,\vx_2,\ldots$ are matrices, and for the sake of notational simplicity we assume that these matrices have been vectorized.
In the case where $n \ll d$, the problem of recovering $\vwstar$ from the measurements is  ill-posed even when $\gstar$ is known. To overcome this, one usually makes additional structural assumptions on the parameters $\vwstar$. Specifically, we assume that the parameters satisfy a notion of ``structural simplicity", which we will now elaborate on.

Suppose we are given a set of {\bf atoms}, $\A = \{ \va \in \R^d \}$, such that any $\vw \in \R^d$ can be written as $\vw = \sum_{\va \in \A} c_a \va$. Although the number of atoms in $\A$ may be uncountably infinite, the sum notation implies that any $\vw$ can be expressed as a linear combination of a finite number of atoms\footnote{This representation need not be unique.}.

%Given a set of atoms $\A$, and a vector $\vw=\sum_{a\in \A}c_a \va$, with $c_a\geq 0$, \cite{venkat} defined the {\bf atomic norm} of a vector $\vw$ as the gauge function \footnote{The function is a norm when the set $\A$ is centrally symmetric, which is a trivial assumption we make in the remainder of this paper} of $\A$:
%\beq
%\label{defanorm}
%\| \vw \|_{\A} = \inf \Bigl\{\sum_{\va} c_a ~\: \vw = \sum_{\va} c_a \va , ~\ c_a \geq 0 ~\ \forall ~\ \va \in \A\Bigr\}
%\eeq

%The atomic norm acts as a convex proxy to the minimum number of atoms needed to represent $\vw$. Indeed, a small atomic norm typically means that the number of atoms used in the representation is small.
Consider the following non convex {\bf atomic cardinality} function:
\beq
\begin{split}
\label{defcard}
\| \vw \|_{\A, 0} = \inf \Bigl\{\sum_{\va} \mathds{1}[c_a > 0 ] ~\ :\vw = \sum_{\va} c_a \va , ~\ c_a \geq 0 ~\ \\
 \forall ~\ \va \in \A\Bigr\}
\end{split}
\eeq
$\mathds{1}[\cdot]$ denotes the indicator function: it is unity when the condition inside the $[\cdot]$ is satisfied, and infinity otherwise. We say that a vector $\vw$ is  ``structurally simple" with respect to an atomic set $\A$ if $\| \vw \|_{\A, 0}$ in \eqref{defcard} is small. The notion of structural simplicity plays a central role in several high dimensional machine learning and signal processing applications:

\begin{enumerate}
\item Sparse regression and classification problems are ubiquitous in several areas, such as neuroscience \cite{ryali2010sparse} and compressed sensing \cite{donoho}. The atoms in this case are merely the signed canonical basis vectors, and the atomic cardinality of a vector $\vw$ is simply the sparsity of $\vw$.
\item The idea of  group sparsity plays a central role in multitask learning \cite{pontil} and computational biology \cite{jacob}, among other applications. The atoms are low dimensional unit disks, and the atomic cardinality of a vector $\vw$ is simply the group sparsity of $\vw$.
\item  Matrix estimation problems that typically appear in problems such as collaborative filtering~\cite{koren} can be modeled as learning vectors with atoms being unit rank matrices and the resulting atomic cardinality being the rank of the matrix.
%\item Integer programming problems look to recover sign vectors~\cite{mangasarian2011probability}, can be modeled with canonical basis vectors as atoms and the atomic cardinality being the sparsity of the vector.
\end{enumerate}

\subsection{Problem Setup: Calibrated loss minimization}
\label{sec:setup}
Our goal in this paper will be to solve an optimization problem of the form
\beq
\label{tosolve}
\hat{g}, \hat{\vw} := \arg \min_{g, \vw} \mathcal{L}(\vy, \mX, \vw, g) + \frac{\lambda}{2} \| \vw \|_2^2 ~\ \textbf{s.t.} ~\ \| \vw \|_{\A, 0} \leq s
\eeq
where $\A$ is a known atomic set, $k$ is a positive integer, and $\mathcal{L}$ is a loss function that is appropraitely designed that we elaborate on next. Notice that in the above formulation we added a squared $\ell_2$ norm penalty to make the objective function strongly convex. In the case when we are dealing with matrix problems we can use the Frobenius norm of $\vw$. The constraint on the atomic cardinality ensures the learning of structurally simple parameters, and indeed makes the problem well posed.

Suppose $\gstar$ was known. Let $\Phistar: \R \rightarrow \R$ be a function such that
$\Phistar' = \gstar$, and consider the following optimization problem.
\begin{align}
\label{def_glm_log_likelihood_single}
\hat{\vw} &:=  \arg \min_{\vw} \frac{1}{n} \sum_{i = 1}^n \Phi_\star(\vw^\top\vx_i) - \vy_i \vw^\top \vx_i+\frac{\lambda}{2} \|\vw\|^2  \notag\\
& \quad \textbf{s.t.}~ \| \vw \|_{\A, 0} \leq s
\end{align}
Modulo the $\| \vw \|_{\A, 0}$ penalty and the regularization terms, the above objective is a sample version of the following stochastic optimization problem:
\begin{equation}
\min_{\vw}\bbE [\Phi_\star(\vw^\top \vx)-y \vw^\top \vx ].
\end{equation}
Since, $\gstar$ is a monotonically increasing function, $\Phistar$ is convex and the above stochastic optimization problem is convex. By taking the first derivative we can verify that the optimal solution satisfies the relation $\bbE[Y|X=x]=\gstar(\vwstar^\top\vx)$. Hence, by defining the loss function in terms of the integral of the transfer function, the loss function is \textit{calibrated} to the transfer function, and automatically adapts to the SIM from which the data is generated. To gain further intuition, notice that when $\gstar$ is linear, then $\Phistar$  is quadratic and the optimization problem in Equation~\eqref{def_glm_log_likelihood_single} is a constrained squared loss minimization problem. When $\gstar$ is logit then the problem in Equation~\eqref{def_glm_log_likelihood_single} is a constrained logistic loss minimization problem.

The optimization problem in Equation~\eqref{def_glm_log_likelihood_single} assumes that we know $\gstar$. When $\gstar$ is unknown, we instead consider the following loss function  in \eqref{tosolve}:
\beq
\label{calibratedloss}
\mathcal{L}(\vy, \mX, \vw, g) := \frac{1}{n}  \sum_{i = 1}^n \Phi(\vw^\top\vx_i) - \vy_i \vw^\top\vx_i
\eeq
where we constrain $\Phi' = g \in \G, $ the set of monotonic, 1-Lipschitz functions.  With this choice of $\mathcal{L}$, our optimization problem becomes
\begin{align}
\label{calibrated-prob}
\hat{g}, \hat{\vw}=&\arg\min_{g\in \cG,\vw} \frac{1}{n}  \sum_{i = 1}^n \Phi(\vw^\top\vx_i) - \vy_i \vw^\top\vx_i + \frac{\lambda}{2} \| \vw \|_2^2 \notag\\
&\text{subject to} ~\ \| \vw \|_{\A, 0} \leq s, \Phi'  = g
\end{align}
Notice that in the above optimization problem we are simultaneously learning a function $g$ as well as a weight vector. This additional layer of complication explains why learning SIMs is a considerably harder problem than learning GLMs where a typical optimization problem is similar to the one in Equation~\eqref{def_glm_log_likelihood_single}. As we will later show in our experimental results this additional complexity in optimization is justified by the excellent results achieved by our algorithm compared to GLM based algorithms such as linear/logistic regression.

\section{The Calibrated Single Index Algorithm}
\label{sec:algo}

Our algorithm to solve the optimizaion problem in
Equation~\eqref{calibrated-prob} is called as Calibrated Single Index algorithm (CSI) and is sketched in Algorithm~\ref{alg:gensim}. CSI interleaves parameter learning via iterative projected gradient
descent and monotonic function learning via the LPAV algorithm.

\subsubsection{ Function learning using LPAV :}
\label{sec:lpav}

We use the LPAV \cite{sim_sham} method to update the function $g$. One way to learn the a monotonic function would be to model the function as a multi-layer neural network and learn the weights of the newtwork using a gradient based algorithm. LPAV is computationally far simpler. Furthermore, learning several parameters of a NN is typically not an option in data-poor settings such as the ones we are interested in. Another alternative is to cast learning $g$ as a dictionary learning problem, which requires a good dictionary at hand, which in turn relies on domain expertise.

%The optimization problem in~\eqref{calibrated-prob} requires us to learn a monotonic, Lipschitz function. One way to do this could be to model this monotonic function as possibly a multiple layer neural network, where each layer has its own activations. Then the optimization problem in~\eqref{calibrated-prob} would reduce to simultaneously estimating the weights of the connections of the neural network and the unknown parameter vector $\vwstar$. Such approaches are indeed valid though will be far more computationally expensive than the LPAV algorithm that we propose here. Another alternative is to assume that we have access to a finite dictionary of monotonic, Lipschitz functions and that the function $g_{\star}$ belongs to the conic hull of this dictionary. With this assumption, one can rewrite the optimization problem in~\eqref{calibrated-prob} into an optimization problem where we simultaneously optimize both for the weights of the dictionary functions and the weight vector $\vwstar$. Such an approach, is indeed attractive when we have a good dictionary in hand. A good dictionary will be heavily application specific and might need excellent domain knowledge. The approach that we take in this paper is to use the LPAV algorithm~\cite{sim_sham}. The LPAV algorithm is computationally far more simpler than neural network learning and does not need any domain knowledge whatsoever. We next explain what the LPAV algorithm does and why is it relevant here.

Given a vector $\vw_{t-1}$, in order to find a function fit $g_t$ that minimizes the objective in ~\eqref{calibrated-prob}, we can look at the first order optimality condition. Differentiating the objective in ~\eqref{calibrated-prob} w.r.t. $\vw$, and assuming that $\lambda \approx 0$ we get $\sum_{i=1}^n (g_t(\vw_{t-1}^\top \vx_i) - y_i )x_i=0$. If, $\bbE[~\vx_i \vx_i ^\top]\propto I_d$, i.e. if we assume that the features are uncorrelated, and the features have similar variance \footnote{The variance assumption can be satisfied by normalizing the features appropriately. Similarly, the uncorrelatedness assumption can be satisfied via a Whitening transformation}, then by elementary algebra we just need the function $g_t$ to optimize the expression $\sum_{i=1}^n (g_t (\vw_{t-1}^\top \vx_i) - y_i)^2$. LPAV solves this exact optimization problem. More
precisely, given data $(p_1,y_1),\ldots (p_n,y_n)$, where
$p_1,\ldots,p_n\in \bbR$ and $p_i = \vw_{t-1}^\top \vx_i$, LPAV outputs a best univariate
monotonic, 1-Lipschitz function $\hat{g}$ that minimizes the squared
error $\sum_{i=1}^n (g(p_i)-y_i)^2$. LPAV does this using the following two step procedure. In the first step, it solves:
\begin{equation}
\begin{aligned}
\label{opt:lpav}
%\begin{aligned}
  \hat{\vz}&=\arg\min_{\vz\in\bbR^n} ~ \|\vz-\vy\|_2^2 \\
  \textbf{s.t.}& ~\ 0 \leq z_j-z_i\leq p_j-p_i ~\text{if~} p_i \leq p_j
  %\end{aligned}
  \end{aligned}
  \end{equation}
  This is a pretty simple convex quadratic programming problem and can be solved using standard methods. In the second step, we define $\hat{g}$ as follows: Let $\hat{g}(p_i)=\hat{z}_i$ for all $i= 1,2,\ldots,n$. To get $\hat{g}$ everywhere else on the real line, LPAV performs linear interpolation as follows: Sort $p_i$  for all $i$ and let  $p_{\bi}$ be the $i^{th}$ entry after sorting. Then, for any $\zeta\in \bbR$, we have
\begin{equation*}
    \hat{g}(\zeta)=
    \begin{cases}
     \hat{z}_{\{1\}}, &  \zeta \leq p_{\{ 1 \}} \\
      \hat{z}_{\{n\} }, &  \zeta \geq p_{\{ n \}} \\
      \mu \hat{z}_{ \{i \} } + (1 - \mu) \hat{z}_{\{i+1\}} &
      \zeta = \mu p_{\bi} + \\
       &(1-\mu) p_{\{ i+1\} }
    \end{cases}
  \end{equation*}
It is easy to see that $\hat{g}$ is a Lipschitz, monotonic function
and attains the smallest least squares error on the given data.

Note that solving the LPAV is not the same as fitting a GLM. Specifically, LPAV finds a function $g()$ that minimizes the squared error between the fitted function and the response.

%Our algorithm \gensim will invoke the LPAV algorithm in each iteration,
%with $p_i=\vw^\top\vx_i$, using an appropriate weight vector $\vw$.

We are now ready to describe CSI.
 \begin{algorithm}
  \begin{algorithmic}[1]
  \REQUIRE  Data: $\mX=[\vx_1,\ldots,\vx_n]\in \bbR^{n\times d}$,
  Labels:  $\vy=[y_1,\ldots,y_n]^\top$,   Iterations: $T>0$, Step
  size: $\eta>0$, parameters $\lambda \geq 0, ~\ s > 0$, atomic set $\A$.
  \STATE Initialize $\vw_0=P_s^{\A}(\mX^\top \vy)$.
  \FOR{t=1,\ldots, T}
  \STATE $g_t \leftarrow LPAV(\mX \vw_{t-1},\vy)$.
 \STATE Calculate $\tilde{\vw}_t \leftarrow \vw_{t-1}-\frac{\eta}{n}\sum_{i=1}^n (g_t(\vw_t^\top \vx_i)-y_i)\vx_i + \lambda \vw_{t-1}$.
  \STATE $\vw_t\leftarrow P_s^{\A} (\tilde{\vw}^t)$
\ENDFOR
  \end{algorithmic}
\caption{\gensim \label{alg:gensim}}
\end{algorithm}
\gensim begins by initializing
$\vw$ to $\vw_0=P^\A_s(\mX^\top \vy)$. Here $P^\A_s(\cdot)$ is a
projection operator that outputs the best $s-$ atomic-sparse representation of
the argument. $s$ is provided as a parameter to \gensim. We then update our estimate of $\gstar$ to
$g_t$ by using the LPAV algorithm on the data projected onto the
vector $\vw_{t-1}$. Using the updated estimate, $g_t$, we update our
weight vector to $\vw_t$ by a single gradient step on the objective function of the optimization problem in
Equation~\eqref{calibrated-prob} with $\Phi=\Phi_{t}$, where $\Phi_t$
and $g_t$ are related by the equation $\Phi_t'=g_t$. This gradient step is followed by an atomic projection step (Step 5 in CSI). While, one can use convergence checks and stopping conditions to decide when to stop, we noticed that few tens of iterations are sufficient, and in our experiments we set $T=50$.

A key point to note is that \gensim is very general: indeed the only step that
depends on the particular structural assumption made is the projection
step (step 5 in Algorithm~\eqref{alg:gensim}).

As long as one can define this projection step $P^\A_s(\cdot)$ for the structural constraint of interest, one can use the CSI algorithm to learn an appropriate high dimensional SIM. As we show next, this projection step is indeed tractable in a whole lot of cases of interest in high dimensional statistics.

Note that the projection can be replaced by a soft thresholding-type operator as well, and the algorithmic performance should be largely unaffected. However, performing hard thresholding is typically more efficient, and has been shown to enjoy comparable performance to soft thresholding operators in several cases.

\subsection{Examples of Atomic Projections}
\label{sec:proj}
A key component of Algorithm \ref{alg:gensim} is the projection operator $P_s^{\A}(\cdot)$, which entirely depends on the atomic set $\A$. Suppose we are given a vector $\vw \in \R^d$, an atomic set $\A$ and a positive integer $s$. Also, let $\vw = \sum_{a \in \A} c_a \va$,
where the $c_a$ achieve the $\inf$ in the sense of \eqref{defcard}. Let $[\vc] \downarrow$ be the elements $c_a$, arranged in descending order by magnitude. We define
\beq
\label{defproj}
P^\A_s(\vw) := \sum_{i = 1}^s ([\vc] \downarrow)_i \va_i
\eeq
where $(\cdot)_i$ is the $i^{th}$ element of the vector, and $\va_i$ denotes the corresponding atom in the original representation. We can see that performing such projections is computationally efficient in most cases:
\begin{itemize}
\item  When the atomic set are the signed canonical basis vectors, the projection is the standard hard thresholding operation: retain the top $s$ magnitude coefficients of $\vw$.
\item Under low rank constraints, $P_s^{\A}(\cdot)$ reduces to retaining the best rank-s approximation of $\vw$. Since $s$ is typically small, this can be done efficiently using power iterations.
\item  When the atoms are low dimensional unit disks, the projection step reduces to computing the norm of $\vw$ restricted to each group, and retaining the top $s$ groups.
\end{itemize}

\subsection{Computational Complexity of \gensim}
To analyze the computational complexity of each iterate of the \gensim
algorithm, we need to analyze the time complexity of the gradient
step, the projection step and the LPAV steps used in \gensim. The
gradient step takes $O(nd)$ time. The projection step for low-rank,
sparse and group sparse cases can be naively implemented using
$O(d\log(d)+s)$ time or via the use of max-heaps in $O(d+s\log(d))$
time. The LPAV algorithm is a quadratic program with immense structure in the
inequality constraints and in the quadratic objective. Using clever algorithmic techniques one can solve this optimization problem in  $O(n \log(n))$ time (See Appendix D in \cite{sim_sham}).  The total runtime complexity for $T$
iterations of \gensim is $O(T(nd+d\log(d)+s+n \log(n)))$, making the algorithm fairly efficient. In most large scale problems, the data is sparse, in which case the $nd$ term can be replaced by $\mbox{nnz}(X)$.

\section{Experimental Results}
\label{sec:expts}

\begin{table*}
\begin{center}
\begin{tabular}{ | l | r | r | r | r | r  | r |r| }
\hline
{\bf dataset} & SLR   & SQH & SLS   & CSI  & Slisotron& SLNN \\
\hline
link ($d = 1840, n=1051$)         &0.976   &   0.946  &   0.908  		&    \textbf{0.981} & 0.959& 0.975 \\
page ($d=3000, n=1051$)       &0.987    &0.912   & 0.941  		& {0.997}& 0.937& \textbf{0.999}\\
ath-rel ($d=17785, n=1427$)   &   0.857  &   0.726 &   0.733		& \textbf{0.879}& 0.826& 0.875\\
aut-mot ($d= 16347, n=1986$)    & 0.916   & 0.837  &  0.796  		& \textbf{0.941}& 0.914& 0.923\\
cryp-ele ($d=22293, n =1975$)     &  0.960  & 0.912  &  0.834 		& {0.990}& 0.910& \textbf{0.994}\\
mac-win ($d=7511, n=1946$)      & 0.636   & 0.615  &  0.639 		& {0.646}& 0.616& \textbf{0.649}\\
\hline
\end{tabular}
\caption{\label{tab:auc}AUC values for various methods on several
  datasets. The entries in bold are the best values.}
\end{center}
\end{table*}
%%%%%%%%%%%%%%%%%%
%\begin{table*}
%\begin{center}
%\begin{tabular}{ | l || r | r | r | r | r |r| r|}
%\hline
%{\bf dataset} & SLR   & SQH & SLS   & CSI & Slisotron& SLNN  \\
%\hline
%link             &0.954    &0.966         &0.946    & \textbf{0.973}& 0.914& 0.939\\
%page           &0.947    &0.947         &0.931   &\textbf{0.977} & 0.925& 0.962\\
%ath-rel        &0.687    &0.696         &0.598    & 0.771	& 0.726& \textbf{0.794}\\
%aut-mot      &0.801    &0.795         &0.542    & \textbf{0.858}& 0.812& 0.855\\
%cryp-ele     &0.888    &0.899         &0.723    & \textbf{0.949}& 0.916& 0.939\\
%mac-win    &0.605    &0.615         &0.630    &\textbf{0.632}& 0.579& 0.590\\
%\hline
%\end{tabular}
%\end{center}
%\caption{\label{tab:acc}Classification accuracy for various methods and on several
%  datasets. The entries in
%  bold are the best values. }
%\end{table*}

We now compare and contrast our method with several other algorithms, in various high dimensional structural settings and on several datasets. We start with the case of standard sparse parameter recovery,  before proceeding to display the effectiveness of our method in multitask/multilabel learning scenarios and also in the structured matrix factorization setting.
\subsection{Sparse Signal Recovery}
We compare our method with several standard algorithms on high dimensional datasets:
\begin{itemize}
\item Sparse classification with the logistic loss { \bf (SLR)} and the squared hinge loss { \bf (SQH)}. We vary the regularization parameter over $\{ 2^{-10}, 2^{-9}, \cdots, 2^9, 2^{10} \}$. We used MATLAB code available in the L1-General library. %\footnote{\url{https://www.cs.ubc.ca/~schmidtm/Software/L1General.html}}.
\item Sparse regression using least squares {\bf SLS}. We used a modified Frank Wolfe method \cite{cogenttsp}, %\footnote{\url{http://www.cs.utexas.edu/~nikhilr/Code.html}},
and varied the regularizer over $\{2^{-5}, 2^{-4}, \cdots, 2^{19}, 2^{20} \}$.
\item Our method {\bf \gensim}. We varied the sparsity of the solution as $\{ d/4, d/8, d/16, \cdots, d/1024 \}$, rounded off to the nearest integer, where $d$ is the dimensionality of the data.
\item \textbf{Slisotron}~\cite{sim_sham} which is an algorithm for learning SIMs in low-dimensions.
\item Single layer feedforward NN (\textbf{SLNN}) trained using Tensorflow \cite{abadi2016tensorflow} and the Adam optimizer used to minimize cross-entropy \cite{kingma2014adam}~\footnote{The settings used are: learning\_rate=0.1, beta1=0.9, beta2=0.999, epsilon=1e-08, use\_locking=False}. We used the early stopping method and validated results over multiple epochs between 50 and 1000, and the number of hidden  units were varied between 5 and 1000. Since, a SLNN is not constrained to fitting a monotonic function, we would expect SLNNs to have smaller bias than SIMs. However, since SLNNs use more parameters, they have larger variance than SIMs.
\end{itemize}
We always perform a $50-25-25$ train-validation-test split of the data, and report the results on the test set.

We tested the algorithms on several datasets: link and page are datasets from the UCI machine learning repository. We also use four datasets from the 20 newsgroups corpus: %\footnote{\url{http://qwone.com/~jason/20Newsgroups/}}
atheism-religion, autos-motorcycle, cryptography-electronics and mac-windows.
We compared the AUC in Table~\eqref{tab:auc} - since several of the datasets are unbalanced - for each of the
methods. The following is a summary:

\begin{itemize}
\item CSI outperforms simple, widely popular learning algorithms such as SLR, SQH, SLS. Often, the difference between CSI and these other algorithms is quite substantial. For example when measuring accuracy, the difference between CSI and either SLR, SQH, SLS on all the datasets is at least 2\% and in many cases as large as $4-5\%$.
\item CSI comfortably outperforms Slisotron on all datasets and often by a margin as large as $5-6\%$. This is expected because Slisotron does not enforce any structure such as sparsity in its updates.
\item The most interesting result is the comparison with SLNN. In spite of its simplicity, we see that \gensim is comparable to and often outperforms SLNN by a slight margin.
\end{itemize}

%xAs one can see from these results the CSI algorithm outperforms all algorithms to obtain the highest accuracy on all but one dataset. A similar story repeats for the AUC metric too where CSI either outperforms all algorithms or is just as good as the best algorithm. Both these results show that the CSI algorithm is an extremely good algorithm for learning flexible high-dimensional statistical models.
%%%%%%%%%%%%%%%%%%%%%%%%%%%%%%%%%%%%%%%%%%%%%%%
%%%%%%%%%%%%%%%%%%%%%%%%%%%%%%%%%%%%%%%%%%%%%%%
%%%%%%%%%%%%%%%%%%%%%%%%%%%%%%%%%%%%%%%%%%%%%%%
%%%%%%%%%%%%%%%%%%%%%%%%%%%%%%%%%%%%%%%%%%%%%%%
%%%%%%%%%%%%%%%%%%%%%%%%%%%%%%%%%%%%%%%%%%%%%%%
%%%%%%%%%%%%%%%%%%%%%%%%%%%%%%%%%%%%%%%%%%%%%%%
%%%%%%%%%%%%%%%%%%%%%%%%%%%%%%%%%%%%%%%%%%%%%%%
%%%%%%%%%%%%%%%%%%%%%%%%%%%%%%%%%%%%%%%%%%%%%%%
%%%%%%%%%%%%%%%%%%%%%%%%%%%%%%%%%%%%%%%%%%%%%%%
%%%%%%%%%%%%%%%%%%%%%%%%%%%%%%%%%%%%%%%%%%%%%%%
%%%%%%%%%%%%%%%%%%%%%%%%%%%%%%%%%%%%%%%%%%%%%%%

\subsection{Group Sparsity: Multilabel and Multitask Learning}
Next, we consider the problem of multi-label learning using group sparsity structure. We consider two datasets.~%\footnote{\url{http://mulan.sourceforge.net/datasets.html}}.
For multilabel learning, the { \bf flags} dataset contains $194$ measurement and $7$ possible labels (based on the colors in the flag). The data is split into $129-65$ measurements, for training and test respectively. Out of the training set, we randomly set aside $10 \%$ of the measurements for validation.

For multitask learning, the {\bf atp7d} dataset consists of $2$ simultaneous regression tasks from $411$ dimensional data with $296$ measurements. We perform a random $80-10-10$ split of the data for training, validation and testing.

We compared our method with group sparse logistic regression and least squares, using the MALSAR package \cite{malsar}. For logistic regression and least squares, the range of parameter values was $\{2^{-10}, 2^{-9}, \cdots, 2^9, 2^{10} \}$. We varied the step size $\eta \in [2^{-6}, 2^2]$ on a log scale for our method, setting the group sparsity parameter to be $5$ for both datasets. Table~\eqref{tab:res_gsparse} shows that our method performs better than both compared methods. For classification, we use the F1 score as a performance measure, since multilabel problems are highly unbalanced, and measures such as accuracy are not indicative of performance.  For multitask learning, we report the MSE.
\begin{table}[H]
\begin{center}
\begin{tabular}{ | l || r | r | r | }
\hline
{\bf dataset}	& Logistic 	& Linear & CSI 	\\
\hline
atp7d (MSE) 	& 1.1257	&  0.8198	&  ${\bm{0.0611}}$		\\
Flags (F1)  	& 0.6458	&  0.5747	&  ${\bm{0.6539}}$		\\
\hline
\end{tabular}
\caption{\label{tab:res_gsparse}Group Sparsity constrained Multitask and Multilabel
  learning. \gensim outperforms both linear and logistic
  regression. The first row reports MSE (lower is better), while the second row is the F1 score (higher is better).}
\end{center}
\end{table}
%%%%%%%%%%%%%%%%%%%%%%%%%%%%%%%%%%%%%%%%%%%%%%%
%%%%%%%%%%%%%%%%%%%%%%%%%%%%%%%%%%%%%%%%%%%%%%%
%%%%%%%%%%%%%%%%%%%%%%%%%%%%%%%%%%%%%%%%%%%%%%%
%%%%%%%%%%%%%%%%%%%%%%%%%%%%%%%%%%%%%%%%%%%%%%%
%%%%%%%%%%%%%%%%%%%%%%%%%%%%%%%%%%%%%%%%%%%%%%%
%%%%%%%%%%%%%%%%%%%%%%%%%%%%%%%%%%%%%%%%%%%%%%%
%%%%%%%%%%%%%%%%%%%%%%%%%%%%%%%%%%%%%%%%%%%%%%%
%%%%%%%%%%%%%%%%%%%%%%%%%%%%%%%%%%%%%%%%%%%%%%%
%%%%%%%%%%%%%%%%%%%%%%%%%%%%%%%%%%%%%%%%%%%%%%%
%%%%%%%%%%%%%%%%%%%%%%%%%%%%%%%%%%%%%%%%%%%%%%%
%%%%%%%%%%%%%%%%%%%%%%%%%%%%%%%%%%%%%%%%%%%%%%%
\subsection{Structured Matrix Factorization}
We now visit the problem of matrix completion in the presence of
graph side information. We consider two datasets, Epinions
and Flixster %\footnote{\url{http://www.librec.net/datasets.html}}.
Both datasets have a (known) social network among the users. We process the
data as follows: we first retain the top 1000 users and items with the
most ratings. Then we sparsify
the data so as to randomly retain only 3000 observations in the
training set, out of which we set aside 300 observations for cross
validation. Furthermore, we binarize the observations at 3, corresponding to ``likes" and ``dislikes" among users and items.\cite{natarajan2015pu} showed that the problem of structured matrix factorization can be cast as the following atomic norm constrained program.
The least squares approach solves the following program:
\beq
\label{ls:grmf}
\min_{W} \frac{1}{N} \| (Y - W)_{\Omega} \|^2_F ~\ \text{s.t.} ~\ \|S_u^{\frac12}U_u^{-1} W U_v^{-T} S_v^{\frac12} \|_* \leq s
\eeq
where $U_u, S_u $ are the singular vectors and singular values of the graph Laplacian of the graph among the rows of $W$ and $U_v, S_v$ are the same for the graph Laplacian corresponding to the graph among columns of $W$.
We use the same atoms in our case, except we replace the loss function by our calibrated loss. We report the MSE in Table~\eqref{tab:grmf}.
\begin{table}[H]
\begin{center}
\begin{tabular}{ | l || r | r | r | r ||}
\hline
{\bf dataset} & $\#$ test set & $\#$ links & LS & \gensim  \\
\hline
Epinions      & 3234    &   61610 & 1.0012    & $\bm{0.9488}$ \\
Flixster      & 64095 &   4016    & 1.0065    & $\bm{0.9823}$ \\
\hline
\end{tabular}
\caption{\label{tab:grmf}Dataset details and performance of different algorithms for
  structured matrix factorization. LS stands for the least squares method in \eqref{ls:grmf}, and CSI is the same method with an unknown nonlinearity.}
\end{center}
\end{table}

\section{Empirical discussion of the convergence of CSI}
\begin{figure}[]
  \centering
    \includegraphics[width=0.3\textwidth]{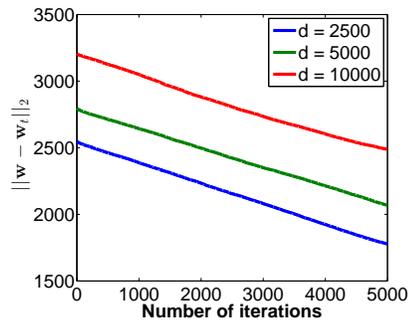}
    \caption{\label{fig:convergence}CSI when applied to a high-dimensional dataset with labels generated by sparse weight vector $\vwstar$ with sparsity level $k=\sqrt{d}$. Each of the three lines show how the $\ell_2$ error decreases with the number of iterations of the CSI algorithm. The covariates $\vx_1,\ldots, \vx_n$ are sampled from $N(0,I_d)$, where $n=500$. CSI is run with $\lambda = 0.001, s=5k$. (best seen in color)}
\end{figure}

 When $\gstar$ is known then the CSI algorithm is basically an iterative gradient descent based algorithm on a convex likelhood function, combined with hard thresholding. Such algorithms have been analyzed and exponential rates of convergence have been established~\cite{agarwal2012fast,jain2014iterative}. These results assume that the likelihood loss function satisfies certain restricted strong convexity and restricted strong smoothness assumptions. This leads to a natural question: Can we establish exponential rates of convergence for the CSI algorithm, for the single index model, i.e. when $\gstar$ is unknown? While, we have been unable to establish a formal analysis of the rates of convergence in this case, we believe that such fast rates might not be achievable in the case of SIM and at best one can achieve much slower sub-linear rates of convergence on the iterates. We support our claim with an experiment, where we study how quickly do the iterates $\vw_1,\vw_2,\ldots$ generated by CSI converge to $\vwstar$ on a synthetic dataset generated using the SIM. Our synthetic experiment is setup as follows: We generate the covariates $\vx_1,\vx_2,\ldots, \vx_n$ from a standard normal distribution $N(0,I_d)$. We use $n =  500$ in our experiment. We then choose $\vwstar\in \bbR^d$ to be $\sqrt{d}$ sparse with the locations of the non-zero entries chosen at random. The non-zero entries are filled with values sampled from $N(0,1)$. Next choose $\gstar$ to be the logistic function $\gstar(\vw^\top \vx)= \frac{2}{1+\exp(-\vw^\top \vx)} - 1$~\footnote{Note that this definition of $\gstar$ is exactly the same as the standard logistic formula $\frac{1}{1+\exp(-\vw^\top\vx)}$. Since we are working with expectations in Equation~\eqref{eq:sim} and not probabilities as is done in classical logistic regression, our formula, on the surface, looks a bit different.} and generate labels in $\{+1, -1\}$ for each $\vx_i$ using a SIM as shown in Equation~\eqref{eq:sim} with the above $\gstar$. For our experiments both $\gstar,\vwstar$ are kept hidden from the CSI algorithm. We run CSI with $\lambda = 0.001$ and $s=5k$. In Figure~\eqref{fig:convergence} we show how the distance of the iterates $\vw_t$ from $\vwstar$ changes as the number of iterations of CSI increases. This result tells us that the distance monotonically decreases with the number of iterations and moreover, the problem is harder as dimensionality increases. Combining the results of~\cite{jain2014iterative} and the simulation result shown in Figure~\eqref{fig:convergence} we make the following conjecture.
\begin{conjecture}
Suppose we are given i.i.d. labeled data which satisfies the SIM $\bbE[y|X = x]=\gstar(\vwstar^\top \vx)$, where $\gstar$ is a $L-$ Lipschitz, monotonic function and $\|\vwstar\|_0 \leq k$. Let $\Phi_{\star}'\defeq\gstar$ be $(\gamma_l, \tau_l)$ restricted strong convex and $(\gamma_u, \tau_u)$ restricted strong smooth, as defined in~\cite{agarwal2012fast}, for the given data distribution. Then with an appropriate choice of the parameters $\lambda$ and $s \geq k$, algorithm CSI with the hard-thresholding operation after $t$ iterations outputs a vector $\vw_t$ that satisfies
\begin{equation}
||\vw_t - \vwstar||_2 \leq f(\gamma_l, \gamma_u) O\left(\frac{Lk\log(d)}{t}\right) +\Delta
\end{equation}
where $\Delta$ is some function that depends on  $k, \log(d), n, \tau_l, \tau_u$ and $f(\gamma_l, \gamma_u)$ is some function dependent on $\gamma_l,\gamma_u$ and is independent of $t$. $\Delta$ represents the statistical error of the iterates that arises due to the presence of limited data.
\end{conjecture}

\section{Conclusions and Discussion}
\label{sec:conc}

In this paper, we introduced \gensim, a unified  algorithm to learn
single index models in high dimensions, under general structural
constraints on the data. The simplicity of our learning algorithm, its versatility, and competitive results makes it a great tool that can be added to a data analyst's toolbox.

%In fact the notion of a critical point
%is also hard to define due to the changing nature of the calibrated
%loss optimization problem. However, experimentally we have
%noticed that the training error keeps decreasing with the number of
%iterations, and perhaps obtaining an upper bound on the training error
%is a more sensible theoretical problem. Either way, this remains a
%very hard and interesting technical problem, and we would like to study \gensim
%theoretically in the future.
\bibliography{simbib}
\bibliographystyle{aaai}

\end{document}